\documentclass[letterpaper]{article} 
\usepackage{aaai2027}
\nocopyright
\usepackage[hyphens]{url}  
\usepackage{graphicx} 
\usepackage{natbib}  
\usepackage{caption} 
\usepackage{algorithm}
\usepackage{algorithmic}
\usepackage{amsmath}
\usepackage{amsmath,amssymb}

\usepackage{newfloat}
\usepackage{listings}
\DeclareCaptionStyle{ruled}{labelfont=normalfont,labelsep=colon,strut=off} 
\floatstyle{ruled}
\newfloat{listing}{tb}{lst}{}
\floatname{listing}{Listing}

\usepackage{booktabs}

\title{ST-WAM: Semantic-Temporal World Action Model for Robust Manipulation under Visual Distribution Shifts}

\author{
    Mingxin Wang\textsuperscript{\rm 1,\rm 2},
    Bin Hu\textsuperscript{\rm 1},
    Bin Qian\textsuperscript{\rm 1},
    Kaitao Jiang\textsuperscript{\rm 2},
    Haoning Wu\textsuperscript{\rm 3},\\
    Feng Yan\textsuperscript{\rm 4},
    Bowen Jing\textsuperscript{\rm 5},
    Ruiyang Hao\textsuperscript{\rm 6},
    Enyi Wang\textsuperscript{\rm 1},
    Kangning Niu\textsuperscript{\rm 2},\\
    Yandan Yang\textsuperscript{\rm 2},
    Mu Xu\textsuperscript{\rm 2},
    Yan Wang\textsuperscript{\rm 1},
    Houde Liu\textsuperscript{\rm 1}\corresponding,
    Tianlun Li\textsuperscript{\rm 2}\corresponding
}

\affiliations{
    \textsuperscript{\rm 1}Tsinghua University,\quad
    \textsuperscript{\rm 2}AMAP-CV-LAB, Alibaba Group\\
    \textsuperscript{\rm 3}Shanghai Jiao Tong University,\quad
    \textsuperscript{\rm 4}Xi'an Jiaotong University\\
    \textsuperscript{\rm 5}The University of Manchester,\quad
    \textsuperscript{\rm 6}King's College London
}

\begin{document}

\maketitle

\begin{abstract}
World Action Models (WAMs) have emerged as a promising paradigm by jointly modeling robot actions and future visual dynamics.
However, their reliance on pixel-generative future supervision can entangle action-relevant state transitions with task-irrelevant visual content, limiting robustness under visual distribution shifts.
We identify \textbf{Training-Distribution Hallucination}, a recurring phenomenon in which futures conditioned on visually shifted observations hallucinate training-domain content rather than remain faithful to the current scene.
A controlled frame-triplet diagnosis further shows that DINOv3 features remain more stable across visual shifts while better preserving task-state distinctions than Wan-VAE latents.
Rather than correcting the predicted futures, we propose \textbf{Semantic-Temporal WAM (ST-WAM)} to improve action robustness by using DINOv3 as a shared semantic representation for future prediction and history retrieval while retaining fine-grained VAE dynamics.
Its Dual-Space Future Experts (DSFE) jointly predict future VAE latents and DINO features, while Current-Anchored Intent Retrieval (CAIR) retrieves task-relevant evidence from recent DINO history under the current visual-language context.
ST-WAM is trained end-to-end without additional embodied pretraining or task-specific annotations, and requires no explicit future generation at inference.
It achieves $98.7\%$ on LIBERO and $92.8\%$ on RoboTwin 2.0; more importantly, compared with Fast-WAM, it improves zero-shot LIBERO-Plus performance by $21.3$ percentage points and more than doubles real-world success under visual shifts from $25.8\%$ to $61.5\%$.
These results demonstrate that semantic-temporal modeling effectively complements pixel-generative dynamics for robust manipulation. The project page is available at \url{https://thu-wangmx.github.io/st-wam/}.
\end{abstract}


\section{Introduction}

\begin{figure}[t]
    \centering
    \includegraphics[width=\columnwidth]{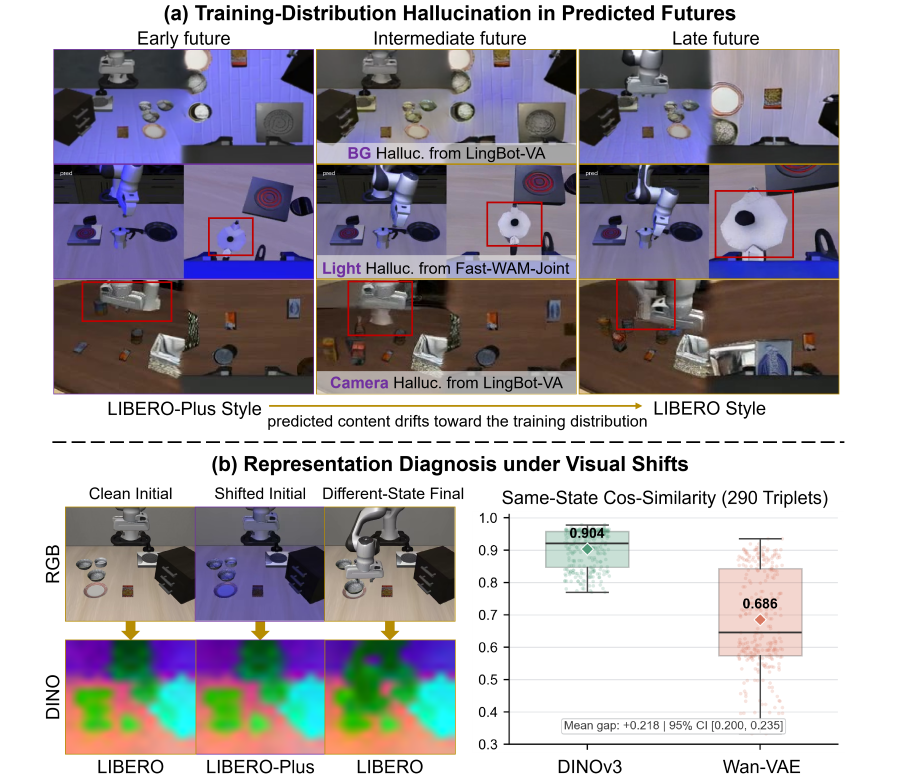}
    \caption{
    (a) Three representative Training-Distribution Hallucination cases from video-generative WAMs under LIBERO-Plus background-texture, illumination, and camera-viewpoint shifts.
    (b) A representative diagnostic triplet and a comparison of
    DINOv3 and Wan-VAE cosine similarities across same-state frames.
    }
    \label{fig:motivation}
    \vspace{-6pt}
\end{figure}

Leveraging world-dynamics priors acquired through large-scale video pretraining, World Action Models (WAMs)~\cite{bi2026motus,kim2026cosmospolicy} jointly model future visual states---typically in VAE latent spaces---and robot actions, providing a promising alternative to VLAs that directly map current observations to actions~\cite{black2025pi0,kim2025openvla,black2025pi05}.
Despite their strong performance on standard manipulation benchmarks, the robustness of video-generative WAMs under visual distribution shifts remains unclear.

To investigate this question, we analyze two representative video-generative WAMs.
As illustrated in Fig.~\ref{fig:motivation}(a), when LingBot-VA~\cite{li2026causalworldmodeling} and Fast-WAM-Joint~\cite{yuan2026fastwam}, both trained only on LIBERO, are evaluated zero-shot on LIBERO-Plus, their predicted videos progressively drift toward LIBERO-style content under perturbations to background textures, illumination, and other scene characteristics.
We refer to this phenomenon as \textbf{Training-Distribution Hallucination}: when the current observation deviates from the training distribution, the predicted future hallucinates training-domain content rather than remaining faithful to the current scene.
To assess its prevalence, we manually audit the predicted futures of both models on 30 randomly sampled cases under each of three visual shifts---background, illumination, and camera viewpoint---totaling 180 predictions; $70.6\%$ distinctly exhibit Training-Distribution Hallucination.
Moreover, Fast-WAM and Fast-WAM-Joint drop from success rates of $97.6\%$ and $98.5\%$ on LIBERO to $51.5\%$ and $59.0\%$ on LIBERO-Plus, respectively~\cite{zhang2026wamrobustness}.
Taken together, the recurring hallucination pattern across two representative models and the substantial zero-shot performance drops reveal a clear robustness limitation of video-generative WAMs under visual distribution shifts.

These observations shift the central question from whether WAMs can predict the future to how the future should be represented for robust control.
Many recent video-generative WAMs model future states in VAE latent spaces~\cite{yuan2026fastwam,li2026causalworldmodeling,ye2026worldactionmodels}.
Optimized primarily for visual reconstruction, these representations
can entangle action-relevant transitions with task-irrelevant or
hallucinated visual content.
Recent studies have explored alternative future representations, including latent action, semantic masks, and spatial value maps~\cite{chen2026lawam,lou2026maskworldmodel,yu2026maskwam,fan2026aim}.
However, existing approaches often rely on large-scale embodied pretraining, multi-stage training, or specialized pipelines for auxiliary supervision.
We instead seek a semantic representation that can be extracted directly from raw observations while remaining stable under visual shifts and discriminative across task states.
Large-scale self-supervised visual encoders, such as
DINOv3~\cite{simeoni2025dinov3}, provide a promising basis through their semantically structured features.

As illustrated in Fig.~\ref{fig:motivation}(b), we conduct a controlled representation diagnosis using 290 frame triplets from LIBERO and LIBERO-Plus.
Each triplet contains two initial frames from the same task with identical robot and object states but different visual conditions, together with a final frame from the same LIBERO demonstration as a different-state reference.
DINOv3 achieves an average cosine similarity of $0.904$ between the same-state initial frames, compared with $0.686$ for Wan-VAE latents.
Moreover, when comparing the shifted initial frame with the other two frames, DINOv3 yields higher similarity to the state-matched clean frame than to the different-state final frame in $95.2\%$ of triplets, versus $60.0\%$ for Wan-VAE.
DINOv3 therefore exhibits both stronger same-state stability under visual shifts and better different-state discriminability than Wan-VAE latents.
Additional details are provided in the supplementary material.

These properties make DINOv3 suitable for two complementary temporal roles: as a future prediction target, it supervises task-relevant state evolution; as a history representation, it provides evidence of recent task progress when visual shifts make current-frame cues unreliable.
Rather than correcting hallucinated future videos, we use these complementary semantic cues to improve action robustness.

In this paper, we propose \textbf{Semantic-Temporal WAM (ST-WAM)}, an end-to-end WAM that uses DINOv3 as a shared semantic representation across two complementary temporal directions.
Prospectively, \textbf{Dual-Space Future Experts (DSFE)} jointly model future VAE latents and DINO features in a three-branch Mixture-of-Transformers with the action expert, coupling fine-grained visual dynamics with visually stable semantic transitions.
Retrospectively, \textbf{Current-Anchored Intent Retrieval (CAIR)} uses the current visual-language context to retrieve task-relevant evidence from recent DINO history for action generation.
ST-WAM requires no additional embodied pretraining, multi-stage training, or task-specific semantic annotations.

Extensive experiments across LIBERO, LIBERO-Plus, RoboTwin 2.0, and five real-world tasks demonstrate the effectiveness of ST-WAM while retaining efficient inference.
Without embodied pretraining, ST-WAM achieves $98.7\%$ on LIBERO and outperforms Fast-WAM by $21.3$ percentage points under zero-shot transfer to LIBERO-Plus.
It further achieves leading performance on RoboTwin 2.0 and more
than doubles real-world success under visual distribution shifts
from $25.8\%$ with Fast-WAM to $61.5\%$.

Our main contributions are summarized as follows:
\begin{itemize}

    \item We identify \textbf{Training-Distribution Hallucination} in video-generative WAMs, where predictions conditioned on visually shifted observations drift toward training-domain content, and provide a controlled frame-triplet diagnosis showing that, compared
    with VAE latents, DINOv3 offers greater same-state stability under visual shifts while better distinguishing different task states.
    \item We propose \textbf{ST-WAM}, which unifies semantic modeling across complementary prospective and retrospective directions: Dual-Space Future Experts (DSFE) jointly model future dynamics in VAE and DINO spaces, while Current-Anchored Intent Retrieval (CAIR) extracts task-relevant evidence from recent DINO history for action generation.
    \item Extensive evaluations in simulation and the real world demonstrate that, without additional embodied pretraining, ST-WAM maintains strong in-distribution performance while substantially improving robustness under visual distribution shifts.
\end{itemize}

\section{Related Work}

\subsection{Vision-Language-Action Models}
VLA models transfer semantic knowledge from pretrained vision-language models to directly map observations and instructions into robot actions~\cite{kim2025openvla,black2025pi0,black2025pi05,physicalintelligence2026pi07,Yan_2025_ICCV,du2026cfvlaefficientcoarsetofineaction}.
Recent methods incorporate temporal context or future prediction: IntentVLA models short-horizon intent from recent history, while DreamVLA, VLA-JEPA, and DeFI introduce predictive objectives into VLA learning~\cite{lian2026intentvla,zhang2025dreamvlavisionlanguageactionmodeldreamed,sun2026vlajepaenhancingvisionlanguageactionmodel,zhang2026disentangledrobotlearningseparate,song2026reconvla}.
In contrast, we study how future and historical information should be represented within WAMs.

\subsection{Video-Generative World Action Models}
World Action Models jointly model future visual states and robot actions.
Representative methods include DreamZero~\cite{ye2026worldactionmodels}, LingBot-VA~\cite{li2026causalworldmodeling}, and Motus~\cite{bi2026motus}, which couple video prediction with action generation.
Fast-WAM~\cite{yuan2026fastwam} and GigaWorld-Policy~\cite{ye2026gigaworld} omit explicit future video generation during deployment to improve inference efficiency.
Despite their different inference designs, their future supervision remains rooted in pixel-generative objectives, potentially entangling action-relevant dynamics with task-irrelevant visual factors.

\begin{figure*}[t]
    \centering
    \includegraphics[width=\textwidth]{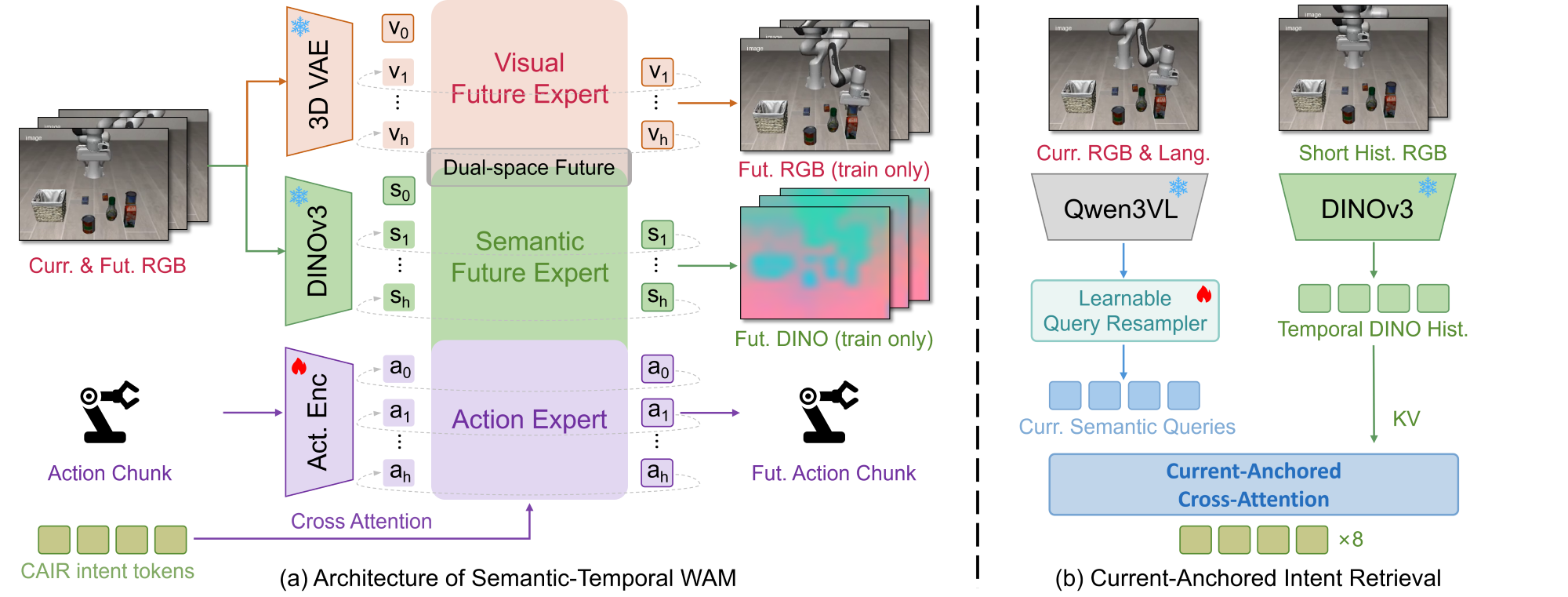}
    \caption{
    Overview of ST-WAM.
    (a): Dual-Space Future Experts (DSFE) jointly model future dynamics in the VAE visual-latent space and the DINOv3 semantic space.
    The visual and semantic future experts interact with the action expert through a three-branch Mixture-of-Transformers.
    (b): Current-Anchored Intent Retrieval (CAIR).
    }
    \label{fig:method_overview}
    \vspace{-6pt}
\end{figure*}

\subsection{WAMs with Alternative Future Representations}
Beyond pixel-generative objectives, recent works explore semantic or spatially structured future representations.
Some methods predict semantic masks~\cite{yu2026maskwam,lou2026maskworldmodel}, while others model geometric-semantic cues, spatial value maps, or compact latent conditions~\cite{fan2026aim,ma2026geosemwam,su2026worldguidance,luo2026beingh07,li2026egowam,liu2026oa,zhang2026learning}.
LDA-1B~\cite{lyu2026lda1b} and LaWAM~\cite{chen2026lawam} model future states directly in DINO feature space, but rely on large-scale embodied pretraining and multi-stage training.
In contrast, ST-WAM retains fine-grained VAE dynamics while incorporating both DINOv3 future supervision and DINO-based history retrieval within an end-to-end WAM, achieving competitive performance without embodied pretraining.

\section{Methodology}

\subsection{Problem Formulation}

Given a current multi-view observation $\mathbf{o}_t$, proprioceptive state $\mathbf{s}_t$, and language instruction $\ell$, the policy predicts an action chunk $\mathbf{a}_{t:t+H-1}$.
We also use a short history $\mathcal{H}_t$ of $M$ preceding observations and a future observation sequence $\mathbf{o}_{t+1:t+K}$.
Let $E_{\mathrm{VAE}}$ and $E_{\mathrm{DINO}}$ denote the frozen Wan2.2 VAE~\cite{wan2025wan} and DINOv3 encoders, respectively.
We encode the observations into two complementary spaces:
\begin{equation}
\mathbf{z}^{v}=E_{\mathrm{VAE}}(\mathbf{o}_{t:t+K}),
\quad
\mathbf{z}^{s}=E_{\mathrm{DINO}}(\mathbf{o}_{t:t+K}).
\label{eq:dual_space_representation}
\end{equation}
For each $r\in\{v,s\}$, we partition $\mathbf{z}^{r}$ into current
conditioning tokens $\mathbf{z}^{r}_{\mathrm{cur}}$ and future prediction
targets $\mathbf{z}^{r}_{\mathrm{fut}}$, i.e.,
$\mathbf{z}^{r}=[\mathbf{z}^{r}_{\mathrm{cur}};
\mathbf{z}^{r}_{\mathrm{fut}}]$.
During training, ST-WAM models the following conditional joint
distribution:
\begin{equation}
p_{\theta}\!\left(
    \mathbf{a}_{t:t+H-1},
    \mathbf{z}^{v}_{\mathrm{fut}},
    \mathbf{z}^{s}_{\mathrm{fut}}
    \,\middle|\,
    \mathbf{o}_{t},
    \mathbf{s}_{t},
    \ell,
    \mathcal{H}_{t}
\right).
\label{eq:problem_formulation}
\end{equation}

At inference, ST-WAM reduces to an action-only policy for efficient deployment:
\begin{equation}
\mathbf{a}_{t:t+H-1}
\sim
\pi_{\theta}\!\left(
    \cdot
    \,\middle|\,
    \mathbf{o}_{t},
    \mathbf{s}_{t},
    \ell,
    \mathcal{H}_{t}
\right).
\label{eq:action_only_inference}
\end{equation}

\subsection{Dual-Space Future Experts}
\label{sec:dsfe}


\paragraph{Unified Visual--Semantic Future Modeling.}
DSFE models future states in complementary VAE-latent and DINO-feature spaces.
The frozen VAE of Wan2.2-TI2V-5B~\cite{wan2025wan} encodes the observation sequence into visual latents $\mathbf{z}^{v}$, while a frozen DINOv3 encoder~\cite{simeoni2025dinov3} extracts dense, frame-wise semantic features $\mathbf{z}^{s}$.
After modality-specific embedding, the pretrained Wan2.2 Video DiT models $\mathbf{z}^{v}_{\mathrm{fut}}$ to preserve the fine-grained visual dynamics inherited from video pretraining, whereas a semantic future DiT models $\mathbf{z}^{s}_{\mathrm{fut}}$ to learn semantic state transitions.

\paragraph{Three-Branch Mixture-of-Transformers.}
As shown in Fig.~\ref{fig:method_overview}(a), the visual and semantic future DiTs, together with the action DiT, form a three-branch Mixture-of-Transformers~\cite{liang2025mot}.
Each branch retains its own parameters and prediction head, while layer-wise mixed attention enables mutual refinement between the two future spaces and allows the action branch to integrate current evidence from both.
During flow-matching training, the three experts jointly denoise future VAE latents, future DINO features, and action tokens, with branch-specific heads estimating flow velocities in their respective spaces.

\begin{figure}[t]
    \centering
    \includegraphics[width=\columnwidth]{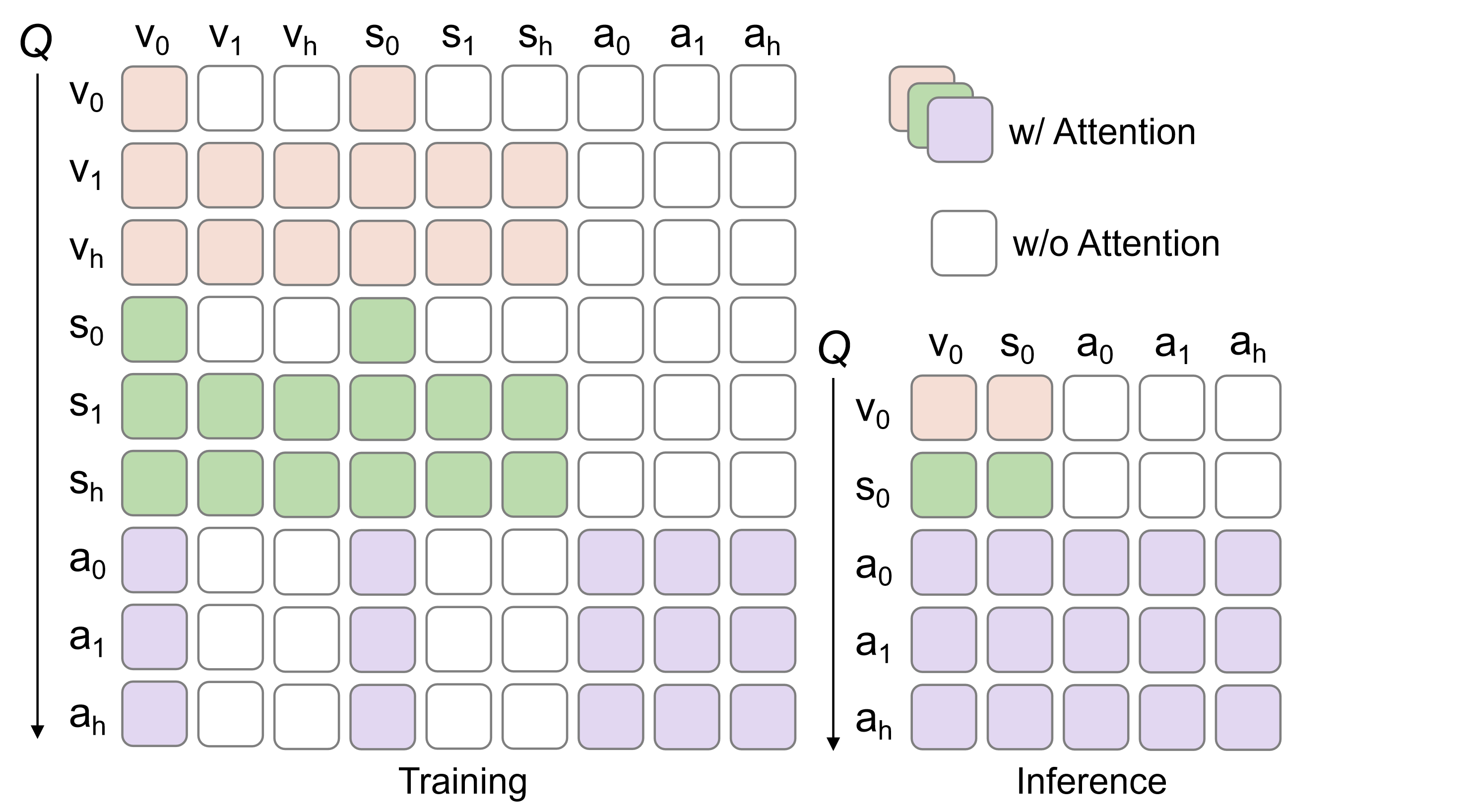}
    \caption{
    Structured cross-branch attention masks during training and inference.
    }
    \label{fig:attention_mask}
    \vspace{-6pt}
\end{figure}

\paragraph{Structured Cross-Branch Attention Mask.}
As illustrated in Fig.~\ref{fig:attention_mask}, we apply an asymmetric mask at each mixed-attention layer.
Clean current VAE and DINO tokens interact within and across their spaces but cannot read future or action tokens, forming leakage-free anchors.
The two noisy future streams attend to both current anchors and each other, enabling mutual refinement across the two future spaces while remaining isolated from action tokens.
Action tokens attend to both current streams and themselves but cannot access either future stream.
This routing prevents future-target leakage into action generation and action-token interference with future modeling, while allowing the future streams to be omitted at inference for efficient deployment.

\subsection{Current-Anchored Intent Retrieval}
\label{sec:cair}


\paragraph{Current-Anchored Semantic Queries.}
As shown in Fig.~\ref{fig:method_overview}(b), given the current observation $\mathbf{o}_t$ and language instruction $\ell$, a frozen Qwen3-VL model extracts its final-layer multimodal hidden states:
\begin{equation}
    \mathbf{H}^{q}_{t}
    =
    E_{\mathrm{Qwen}}(\mathbf{o}_{t},\ell)
    \in
    \mathbb{R}^{B\times L_q\times d_q},
\end{equation}
where $L_q$ is the sequence length and $d_q$ is the hidden dimension of Qwen3-VL.
A bank of $N_I$ learnable queries $\mathbf{Q}$, each with dimension $d_r$, attends to the Qwen3-VL features to produce a compact set of current semantic tokens:
\begin{equation}
    \mathbf{U}^{0}_{t}
    =
    \mathrm{MHA}
    \left(
        \mathbf{Q},
        P_q\mathbf{H}^{q}_{t},
        P_q\mathbf{H}^{q}_{t}
    \right).
\end{equation}
Here, $\mathbf{Q}$ serves as the query, while the projected Qwen3-VL features serve as keys and values.
$P_q$ is a learnable linear projection from the Qwen3-VL hidden dimension $d_q$ to the hidden dimension $d_r$, and $\mathbf{U}^{0}_{t}\in\mathbb{R}^{B\times N_I\times d_r}$ denotes the current semantic tokens before history fusion.
These tokens jointly encode the current scene and instruction, serving as semantic anchors that guide subsequent intent retrieval.

\paragraph{Semantic History Retrieval.}
For each observation in the short history $\mathcal{H}_t$, the frozen DINOv3 encoder extracts dense patch features, providing a visually stable semantic representation from which task-relevant historical evidence can be retrieved.
The features from all $M$ observations are linearly projected, augmented with learnable temporal embeddings, and concatenated into a history-token sequence $\mathbf{R}_t$.

Starting from $\mathbf{U}^{0}_{t}$, CAIR applies $L$ cross-attention blocks, using the current semantic tokens as queries and $\mathbf{R}_t$ as keys and values.
This current-conditioned interaction selectively retrieves historical evidence relevant to the present task state while reducing sensitivity to task-irrelevant visual variations.
After $L$ blocks, the refined tokens  $\mathbf{U}^{L}_{t}$ are projected into the action-context space:
\begin{equation}
    \mathbf{I}_{t}
    =
    P_o\left(\mathbf{U}^{L}_{t}\right)
    \in
    \mathbb{R}^{B\times N_I\times d_c},
\end{equation}
where $P_o$ is the output projection and $d_c$ is the context dimension of the action expert.
We refer to $\mathbf{I}_t$ as short-horizon intent tokens: a latent, label-free summary of recent task progress relevant to current action decision, rather than an explicitly supervised variable.

For expert conditioning, the frozen T5 encoder maps the language instruction into the shared language context
$\mathbf{C}_{\ell}=E_{\mathrm{T5}}(\ell)$.
After appending the projected proprioceptive token $P_p(\mathbf{s}_t)$, where $P_p$ is a learnable projection, the contexts of the three experts are
\begin{equation}
    \mathbf{C}_{v}
    =
    \mathbf{C}_{s}
    =
    [\mathbf{C}_{\ell}; P_p(\mathbf{s}_t)],
    \qquad
    \mathbf{C}_{a}
    =
    [\mathbf{C}_{\ell}; P_p(\mathbf{s}_t); \mathbf{I}_t].
\end{equation}
Each expert receives its corresponding context through cross-attention at every DiT block. 
Thus, the short-horizon intent tokens are injected only into the action expert and optimized end-to-end through action flow matching, allowing semantic evidence from history to guide action generation without altering the conditioning contexts of the DSFE branches.

Together, DSFE and CAIR use DINOv3 in complementary temporal
directions: DSFE prospectively supervises future semantic dynamics, whereas CAIR retrospectively retrieves task-relevant evidence from recent semantic history for robust action generation under visual distribution shifts.

\subsection{Joint Flow-Matching Objective}
\label{sec:flow_matching}

We jointly train the visual future, semantic future, and action experts using flow matching~\cite{lipman2022flow}.
Let the clean branch targets be
$\mathbf{x}^{v}=\mathbf{z}^{v}_{\mathrm{fut}}$,
$\mathbf{x}^{s}=\mathbf{z}^{s}_{\mathrm{fut}}$, and
$\mathbf{x}^{a}=\mathbf{a}_{t:t+H-1}$.
For each branch $r\in\{v,s,a\}$, we sample a timestep $\tau_r\in[0,1]$ and Gaussian noise
$\boldsymbol{\epsilon}^{r}\sim\mathcal{N}(\mathbf{0},\mathbf{I})$.
The noisy input is constructed through linear interpolation:
\begin{equation}
    \mathbf{x}^{r}_{\tau_r}
    =
    (1-\tau_r)\mathbf{x}^{r}
    +
    \tau_r\boldsymbol{\epsilon}^{r}.
\end{equation}
Here, $\tau_r=0$ corresponds to the clean target and $\tau_r=1$ to pure noise.
The corresponding target velocity is
\begin{equation}
    \mathbf{u}^{r}
    =
    \boldsymbol{\epsilon}^{r}-\mathbf{x}^{r}.
\end{equation}

Because the two future experts exchange information through mixed attention, we set
$\tau_v=\tau_s=\tau_f$ to synchronize their denoising stages, while sampling $\tau_a$ independently from the same timestep distribution.
Gaussian noises remain independent across all three branches, and action denoising is therefore decoupled from the two training-only future flows.

The clean current tokens, noisy future tokens, and noisy action tokens are processed in a single MoT forward pass under the structured attention mask, producing branch-specific velocity estimates
$\widehat{\mathbf{u}}_{\theta}^{r}$.
For each branch, we optimize
\begin{equation}
    \mathcal{L}_{r}
    =
    \mathbb{E}
    \left[
        w(\tau_r)
        \left\|
            \widehat{\mathbf{u}}_{\theta}^{r}
            -
            \mathbf{u}^{r}
        \right\|_{2}^{2}
    \right],
    \qquad
    r\in\{v,s,a\},
\end{equation}
where $w(\tau_r)$ denotes the timestep-dependent weighting of the flow scheduler.
The overall training objective is
\begin{equation}
    \mathcal{L}
    =
    \lambda_{v}\mathcal{L}_{v}
    +
    \lambda_{s}\mathcal{L}_{s}
    +
    \lambda_{a}\mathcal{L}_{a},
\end{equation}
where $\mathcal{L}_{v}$, $\mathcal{L}_{s}$, and $\mathcal{L}_{a}$ denote the visual, semantic, and action expert losses, respectively, and $\lambda_{v}$, $\lambda_{s}$, and $\lambda_{a}$ are their corresponding loss weights.

\section{Experiments}

\subsection{Experimental Setup}
\label{sec:experimental_setup}

\paragraph{Benchmarks and Protocols.}
We evaluate in-distribution manipulation performance on the four LIBERO suites~\cite{liu2023libero}: Spatial, Object, Goal, and Long, covering 40 tasks with 50 evaluation rollouts per task.
For out-of-distribution evaluation, we directly evaluate the LIBERO-trained policy on LIBERO-Plus~\cite{fei2026liberoplus} without fine-tuning, which comprises 10,030 test cases spanning seven perturbation dimensions.
We additionally evaluate bimanual manipulation on RoboTwin 2.0~\cite{chen2025robotwin20} and ST-WAM is trained on a mixture of 2,500 clean and 25,000 heavily randomized demonstrations and each task is evaluated over 100 trials in both clean and randomized settings.

\paragraph{Baselines.}
We compare ST-WAM against a broad range of representative methods.
These include VLAs such as $\pi_0$~\cite{black2025pi0}, $\pi_{0.5}$~\cite{black2025pi05}; video-generative WAMs such as Fast-WAM~\cite{yuan2026fastwam}, Motus~\cite{bi2026motus}, and LingBot-VA~\cite{li2026causalworldmodeling}; and methods exploring alternative future representations, such as LaWAM~\cite{chen2026lawam}, Mask World Model~\cite{lou2026maskworldmodel}, MaskWAM~\cite{yu2026maskwam}.

\paragraph{Real-World Evaluation.}

We evaluate ST-WAM on an Agilex Piper 6-DoF single-arm robot using 50 demonstrations per task under fixed nominal visual conditions. We consider five tasks with diverse temporal and geometric requirements: (1) \emph{Arrange Flowers}, inserting three bouquets into a vase; (2) \emph{Drawer Organization}, opening a drawer, placing a pen inside, and closing it; (3) \emph{Bean Scooping}, transferring beans from a plate to a bowl and returning the spoon; (4) \emph{Arrange Fruits}, placing five fruits into a basket; and (5) \emph{Hang Mug}, hanging a mug on a rack. 
We compare ST-WAM with $\pi_0$~\cite{black2025pi0} and Fast-WAM~\cite{yuan2026fastwam}, with all methods post-trained separately for each task using the same demonstrations.
We test them under nominal and four visual-shift conditions without further fine-tuning:
\emph{Background}, which replaces the tabletop texture with unseen patterns;
\emph{Lighting}, which changes the illumination intensity;
\emph{Object Appearance}, which changes object colors or instances while preserving their geometry and task function; and
\emph{Compound}, which applies all three shifts simultaneously.
Each method is evaluated over 30 trials per task and condition using predefined object initializations to ensure fair comparison.

\paragraph{Implementation Details.}
The visual future expert uses the 5B Video DiT of Wan2.2-TI2V-5B~\cite{wan2025wan}, while the semantic and action experts use 1B DiTs initialized from the pretrained Wan2.2 weights.
The Wan2.2 VAE and T5 encoder, DINOv3 ViT-S/16 visual encoder~\cite{simeoni2025dinov3}, and Qwen3-VL-4B-Instruct~\cite{bai2025qwen3vl} remain frozen throughout training.

We set the action horizon to $H=32$ and the future horizon to $K=8$, with future observations sampled every four control steps.
For CAIR, we set $M=4$ and construct $\mathcal{H}_t$ from observations at frame indices $\{t-24, t-16, t-8, t-1\}$.
The retrieval uses $L=2$ cross-attention blocks and $N_I=8$ learnable queries, producing eight intent tokens.

We train on LIBERO and RoboTwin 2.0 for 10 and 5 epochs with global batch sizes of 128 and 1,024, respectively.
We use AdamW ($\mathrm{lr}=1\times10^{-4}$, weight decay $0.01$), cosine decay, BF16 precision, gradient clipping at $1.0$, and a shifted flow-matching schedule with shift $5.0$; the loss weights are $(\lambda_v,\lambda_s,\lambda_a)=(1.0,0.02,1.0)$.
At inference, we use 10 flow-integration steps and execute 10 actions before replanning; additional details are provided in the supplementary material.

\subsection{Main Results}
\label{sec:main_results}

\begin{table}[t]
\centering
\setlength{\tabcolsep}{2.0pt}
\renewcommand{\arraystretch}{1.08}
\resizebox{\columnwidth}{!}{
\begin{tabular}{lcccccc}
\toprule
Method
& \shortstack{Emb. PT.}
& Spat.
& Obj.
& Goal
& Long
& Avg. \\
\midrule

\multicolumn{7}{c}{\textit{Vision-Language-Action Models}} \\
\midrule
$\pi_0$~\cite{black2025pi0}
    & Yes & 98.0 & 96.8 & 94.4 & 88.4 & 94.4 \\
$\pi_{0.5}$~\cite{black2025pi05}
    & Yes & 98.8 & 98.2 & 98.0 & 92.4 & 96.9 \\
IntentVLA~\cite{lian2026intentvla}
    & No & 99.3 & 99.7 & 98.1 & 97.4 & 98.6 \\

\midrule
\multicolumn{7}{c}{\textit{Video-Generative World Action Models}} \\
\midrule
Fast-WAM~\cite{yuan2026fastwam}
    & No & 98.2 & \textbf{100.0} & 97.0 & 95.2 & 97.6 \\
Motus~\cite{bi2026motus}
    & Yes & 96.8 & 99.8 & 96.6 & 97.6 & 97.7 \\
LingBot-VA~\cite{li2026causalworldmodeling}
    & Yes & 98.5 & 99.6 & 97.2 & \textbf{98.5} & 98.5 \\

\midrule
\multicolumn{7}{c}{\textit{WAMs with Alternative Future Representations}} \\
\midrule
Mask World Model~\cite{lou2026maskworldmodel}
    & No & 98.8 & \textbf{100.0} & 98.2 & 96.0 & 98.3 \\
MaskWAM~\cite{yu2026maskwam}
    & No & 98.8 & \textbf{100.0} & 98.2 & 96.4 & 98.4 \\
GeoSem-WAM~\cite{ma2026geosemwam}
    & No & 99.0 & \textbf{100.0} & 98.2 & 97.0 & 98.6 \\
LaWAM~\cite{chen2026lawam}
    & Yes & \textbf{99.4} & 99.6 & 98.4 & 97.0 & 98.6 \\

\midrule
\textbf{ST-WAM (Ours)}
    & No & 99.0 & \textbf{100.0} & \textbf{99.0} & 96.8
    & \textbf{98.7} \\
\bottomrule
\end{tabular}
}
\caption{
Success rates (\%) on the four LIBERO suites.
\textbf{Emb. PT.} indicates large-scale pretraining on robot
trajectories or embodied videos before LIBERO adaptation.
}
\label{tab:libero_results}
\end{table}

\begin{table}[t]
\centering
\setlength{\tabcolsep}{4.0pt}
\renewcommand{\arraystretch}{1.08}
\resizebox{\columnwidth}{!}{
\begin{tabular}{lcccc}
\toprule
Method
& Emb. PT.
& Clean
& Random
& Avg. \\
\midrule

$\pi_0$~\cite{black2025pi0}
& Yes & 65.92 & 58.40 & 62.16 \\

$\pi_{0.5}$~\cite{black2025pi05}
& Yes & 82.74 & 76.76 & 79.75 \\

GigaWorld-Policy~\cite{ye2026gigaworld}
& Yes & 87.00 & 85.00 & 86.00 \\

Motus~\cite{bi2026motus}
& Yes & 88.66 & 87.02 & 87.84 \\

LaWAM~\cite{chen2026lawam}
& Yes & 92.64 & 89.80 & 91.22 \\

Fast-WAM~\cite{yuan2026fastwam}
& No & 91.88 & 91.78 & 91.83 \\

LingBot-VA~\cite{li2026causalworldmodeling}
& Yes & 92.90 & 91.50 & 92.20 \\

GeoSem-WAM~\cite{ma2026geosemwam}
& No & 92.94 & 92.14 & 92.54 \\

\midrule
\textbf{ST-WAM (Ours)}
& No
& \textbf{93.06}
& \textbf{92.48}
& \textbf{92.77} \\

\bottomrule
\end{tabular}
}
\caption{
Success rates (\%) on RoboTwin 2.0 under the standard mixed
clean-and-randomized training setting.
}
\label{tab:robotwin_results}
\vspace{-6pt}
\end{table}

\begin{table*}[t]
\centering
\small
\setlength{\tabcolsep}{4.0pt}
\renewcommand{\arraystretch}{1.08}
\resizebox{0.96\textwidth}{!}{%
\begin{tabular}{lccccccccc}
\toprule
Method
& \shortstack{Emb. PT.}
& Camera
& Robot
& Lang.
& Light
& BG
& Noise
& Layout
& Overall \\
\midrule

UniVLA~\cite{bu2025univla}
& Yes & 1.8 & 46.2 & 69.6 & 69.0 & 81.0 & 21.2 & 31.9 & 42.9 \\

$\pi_0$~\cite{black2025pi0}
& Yes & 13.8 & 6.0 & 58.8 & 85.0 & 81.4 & 79.0 & 68.9 & 53.6 \\

$\pi_0$-FAST~\cite{pertsch2025fast}
& Yes & \textbf{65.1} & 21.6 & 61.0 & 73.2 & 73.2 & 74.4 & 68.8 & 61.6 \\

RIPT-VLA (OFT)~\cite{tan2026riptvla}
& Yes & 55.2 & 31.2 & 77.6 & 88.4 & 91.6 & 73.5 & 74.2 & 68.4 \\

OpenVLA-OFT~\cite{kim2025openvlaoft}
& Yes & 56.4 & 31.9 & 79.5 & 88.7 & 93.3 & 75.8 & 74.2 & 69.6 \\

X-VLA~\cite{zheng2026xvla}
& Yes & 23.4 & \textbf{89.7} & 75.7 & 88.2 & \textbf{96.0} & 62.7 & 71.8 & 71.4 \\

Fast-WAM~\cite{yuan2026fastwam}
& No & 16.4 & 44.5 & 68.9 & 78.2 & 53.7 & 37.7 & 60.7 & 51.5 \\

Fast-WAM-Joint~\cite{yuan2026fastwam}
& No & 34.0 & 55.1 & \textbf{88.9} & 90.0 & 44.9 & 33.3 & 73.4 & 59.0 \\

\midrule
\textbf{ST-WAM (Ours)}
& No
& 55.4
& 60.1
& 79.3
& \textbf{93.0}
& 74.2
& \textbf{79.5}
& \textbf{74.3}
& \textbf{72.8} \\
\bottomrule
\end{tabular}%
}
\caption{
Zero-shot success rates (\%) on LIBERO-Plus.
Baseline results from ~\cite{zhang2026wamrobustness,chen2026abot}.
}
\label{tab:libero_plus}
\vspace{-8pt}
\end{table*}

\paragraph{Performance on LIBERO.}
As shown in Table~\ref{tab:libero_results}, ST-WAM achieves an average success rate of $98.7\%$, the highest among the compared methods.
Without embodied pretraining, it outperforms recent strong WAMs such as Motus ($97.7\%$)~\cite{bi2026motus} and LingBot-VA ($98.5\%$)~\cite{li2026causalworldmodeling}.
Thus, ST-WAM further improves in-distribution performance.

\paragraph{Performance on RoboTwin 2.0.}
As shown in Table~\ref{tab:robotwin_results}, ST-WAM achieves success rates of $93.06\%$ and $92.48\%$ in the clean and randomized settings, respectively, yielding the highest average success rate of $92.77\%$ among the compared methods.
Without embodied pretraining, ST-WAM outperforms both the embodied-pretrained LingBot-VA~\cite{li2026causalworldmodeling} ($92.20\%$) and the closely related Fast-WAM~\cite{yuan2026fastwam} ($91.83\%$).
These consistent results demonstrate the effectiveness of ST-WAM for bimanual manipulation across clean and randomized environments.

\paragraph{Zero-Shot Generalization on LIBERO-Plus.}
As shown in Table~\ref{tab:libero_plus}, ST-WAM achieves an overall success rate of $72.8\%$.
Without embodied pretraining, it surpasses several embodied-pretrained VLAs, including OpenVLA-OFT ($69.6\%$), RIPT-VLA ($68.4\%$), and X-VLA ($71.4\%$).
More importantly, ST-WAM improves the closely matched Fast-WAM baseline~\cite{yuan2026fastwam} from $51.5\%$ to $72.8\%$, a gain of $21.3$ percentage points.
This improvement is consistent across all seven perturbation categories, with particularly large gains of $39.0$ and $41.8$ percentage points under camera and sensor-noise perturbations, respectively.
It also outperforms Fast-WAM-Joint on all six non-language perturbations.
Together, these results suggest that complementing fine-grained VAE dynamics with DINO-based semantic supervision reduces reliance on low-level visual cues and substantially improves out-of-distribution robustness.

\paragraph{Inference Efficiency.}
On RoboTwin 2.0, we benchmark the complete action-chunk inference call, including all model components active at deployment, on a single NVIDIA A100-80GB GPU using BF16 precision and 10 flow-integration steps.
Averaged over 20 synchronized runs, ST-WAM generates a 32-step action chunk in $756.17$ ms, compared with $609.30$ ms for Fast-WAM~\cite{yuan2026fastwam}.
This $1.24\times$ latency represents a moderate overhead for improved robustness while retaining sub-second inference.

\begin{figure}[t]
    \centering
    \includegraphics[width=\columnwidth]{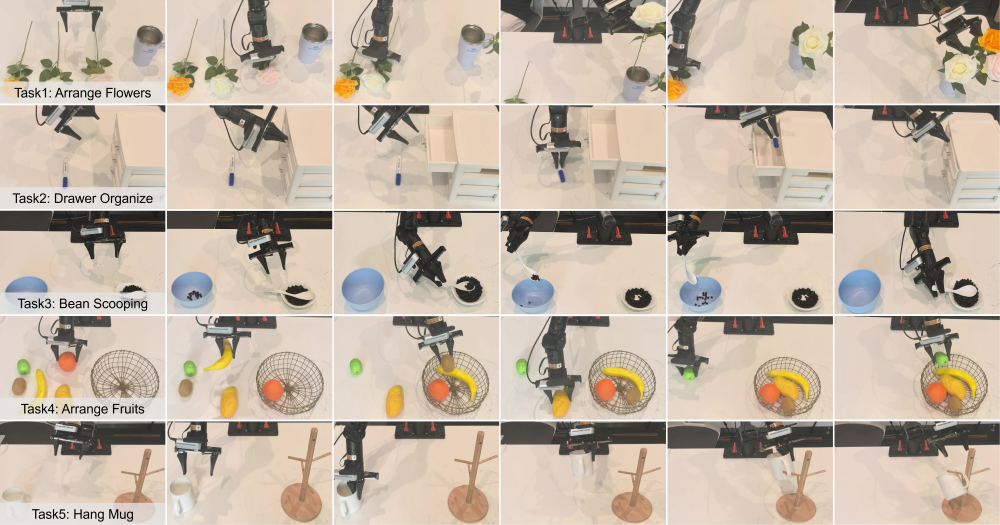}
    \caption{Real-world evaluation of ST-WAM on five tasks with diverse temporal and geometric requirements.}
    \label{fig:real_world_experiments}
    \vspace{-6pt}
\end{figure}

\paragraph{Real-World Generalization.}
As shown in Fig.~\ref{fig:real_world_experiments} and Table~\ref{tab:real_world_results}, ST-WAM achieves $79.3\%$ average success under the nominal condition, outperforming Fast-WAM and $\pi_0$ by $14.6$ and $32.0$ percentage points, respectively.
Under visual distribution shifts, ST-WAM achieves $61.5\%$, surpassing Fast-WAM and $\pi_0$ by $35.7$ and $28.7$ points,
respectively, and retaining a substantial advantage under the compound shift ($48.0\%$ vs.\ $15.3\%$).
Notably, Fast-WAM drops by $38.9$ points from the nominal to shifted conditions, compared with only $17.8$ points for ST-WAM, suggesting that pixel-generative future representations are particularly sensitive to visual distribution shifts.
Removing the semantic future expert or CAIR reduces shifted-condition
performance to $41.0\%$ and $43.7\%$, respectively, confirming their complementary contributions to real-world robustness.
Task-wise results under visual shifts are provided in the supplementary material.

\begin{table*}[t]
\centering
{
\small
\setlength{\tabcolsep}{1.9pt}
\renewcommand{\arraystretch}{1.10}
\begin{tabular*}{\textwidth}{
@{\extracolsep{\fill}}l*{11}{c}@{}
}
\toprule
& \multicolumn{6}{c}{Nominal Environment}
& \multicolumn{5}{c}{Visual Distribution Shifts} \\
\cmidrule(lr){2-7}
\cmidrule(lr){8-12}
Method
& Flower
& Drawer
& Scoop
& Fruit
& Hang
& Avg.
& BG
& Light
& Obj. App.
& Comp.
& Avg. \\
\midrule

$\pi_0$~\cite{black2025pi0}
& 56.7
& 46.7
& 30.0
& 46.7
& 56.7
& 47.3
& 33.3
& 40.7
& 35.3
& 22.0
& 32.8 \\

Fast-WAM~\cite{yuan2026fastwam}
& 70.0
& 66.7
& 46.7
& 66.7
& 73.3
& 64.7
& 27.3
& 35.3
& 25.3
& 15.3
& 25.8 \\

\midrule

w/o Semantic Future Expert
& 76.7
& 73.3
& 56.7
& 73.3
& 80.0
& 72.0
& 43.3
& 50.0
& 41.3
& 29.3
& 41.0 \\

w/o CAIR
& 80.0
& 76.7
& 60.0
& 76.7
& 83.3
& 75.3
& 46.0
& 52.7
& 44.0
& 32.0
& 43.7 \\

\midrule

\textbf{ST-WAM (Ours)}
& \textbf{86.7}
& \textbf{80.0}
& \textbf{66.7}
& \textbf{76.7}
& \textbf{86.7}
& \textbf{79.3}
& \textbf{66.0}
& \textbf{70.0}
& \textbf{62.0}
& \textbf{48.0}
& \textbf{61.5} \\

\bottomrule
\end{tabular*}
}
\caption{
Real-world success rates (\%) under the nominal condition and
visual distribution shifts.
Results for each visual shift are averaged across all five tasks.
``Comp.'' combines background, lighting, and object-appearance
shifts.
}
\label{tab:real_world_results}
\end{table*}

\subsection{Ablation and Qualitative Analysis}
\label{sec:ablation}

\begin{table*}[t]
\centering
{
\small
\setlength{\tabcolsep}{5.5pt}
\renewcommand{\arraystretch}{1.08}
\begin{tabular}{lcccc}
\toprule
Variant
& Future Prediction
& Intent Conditioning
& LIBERO
& LIBERO-Plus \\
\midrule

Fast-WAM~\cite{yuan2026fastwam}
& VAE
& None
& 97.6
& 51.5 \\

DINO Future Only
& DINO
& None
& 96.3
& 39.7 \\

Dual-Space w/o CAIR
& VAE + DINO
& None
& 97.8
& 66.4 \\

\midrule

w/o Semantic Future Expert
& VAE
& CAIR (DINO history)
& 97.3
& 63.5 \\

Semantic Expert w/o Future Obj.
& VAE
& CAIR (DINO history)
& 95.8
& 62.9 \\

\midrule

Naive History Retrieval
& VAE + DINO
& Unanchored DINO history
& 96.3
& 56.5 \\

Qwen Current Only
& VAE + DINO
& Qwen current
& 96.5
& 62.3 \\

CAIR with VAE History
& VAE + DINO
& CAIR (VAE history)
& 96.3
& 64.7 \\

\midrule

\textbf{ST-WAM (Ours)}
& VAE + DINO
& \textbf{CAIR (DINO history)}
& \textbf{98.7}
& \textbf{72.8} \\

\bottomrule
\end{tabular}
}
\caption{
Ablation results on LIBERO and LIBERO-Plus.
``Intent Conditioning'' denotes the alternative action-expert
contexts used to ablate the design choices of CAIR.
}
\label{tab:ablation}
\vspace{-10pt}
\end{table*}

Beyond the component-level real-world ablations, we conduct finer-grained controlled studies on LIBERO and LIBERO-Plus, as shown in Table~\ref{tab:ablation}, to disentangle the design choices underlying DSFE and CAIR. Specifically, we address three critical questions:

\noindent\textbf{Q1: Are VAE and DINO future representations complementary or interchangeable?}
We first compare different future representation spaces without intent conditioning.
The \emph{DINO Future Only} variant achieves $39.7\%$ on LIBERO-Plus,
below the $51.5\%$ of the VAE-based Fast-WAM.
In contrast, jointly modeling future states in the VAE and DINO spaces improves the success rate to $66.4\%$ even without CAIR.
These results indicate that DINO semantics cannot directly replace VAE dynamics: VAE latents preserve fine-grained visual and motion information, whereas DINO features provide complementary object- and state-level semantics.

\noindent\textbf{Q2: Is explicit semantic future prediction necessary?}
Keeping CAIR fixed, \emph{w/o Semantic Future Expert} removes
the semantic stream and achieves $63.5\%$ on LIBERO-Plus.
The parameter-matched \emph{Semantic Expert w/o Future Obj.}
retains the semantic DiT, current-DINO conditioning, and
mixed-attention interactions, but removes its future target and loss, yielding $62.9\%$.
In contrast, the full model reaches $72.8\%$, showing that the
gain cannot be explained by current DINO features or additional model capacity alone, but requires explicit future-semantic prediction.
\begin{figure}[!ht]
    \centering
    \includegraphics[width=\columnwidth]{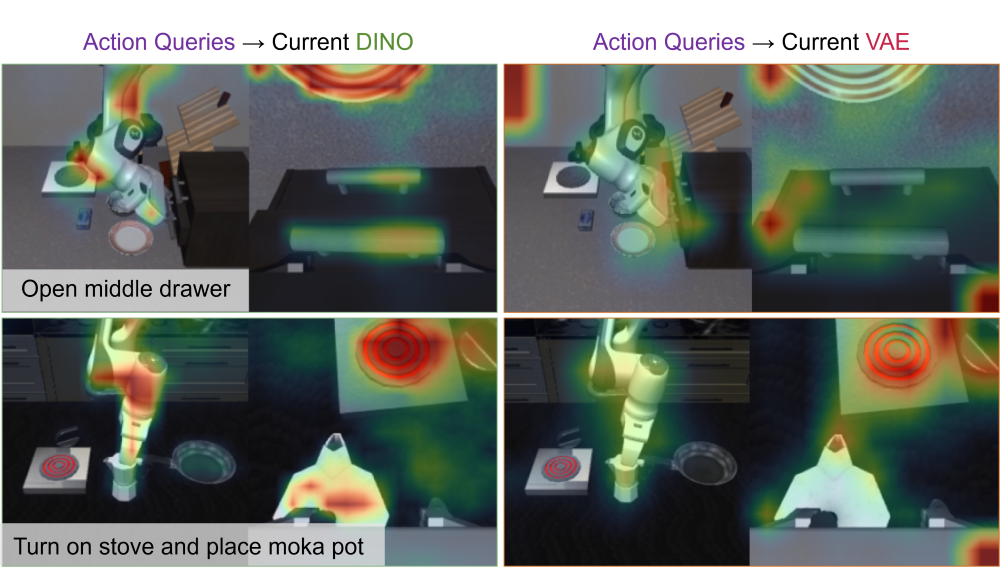}
    \caption{
    Attention heatmaps on two representative tasks.
    Warmer colors indicate stronger relative attention within each map.
    }
    \label{fig:cross_branch_attention}
    \vspace{-10pt}
\end{figure}

\noindent\textbf{Q3: How should short-horizon history be represented and retrieved?}
The \emph{Naive History Retrieval} variant compresses DINO history with unanchored learnable queries and achieves only $56.5\%$ on LIBERO-Plus, versus $72.8\%$ for the full model, demonstrating the necessity of a current semantic anchor.
Removing history entirely in \emph{Qwen Current Only} yields $62.3\%$, ruling out VLM-derived current-frame semantics alone as the source of improvement.
Replacing DINO history with Wan-VAE latents in \emph{CAIR with VAE History}, while retaining the same Qwen anchor and retrieval architecture, obtains $64.7\%$.
All three alternative conditioning schemes underperform the \emph{Dual-Space w/o CAIR} baseline ($66.4\%$), indicating that improperly represented or retrieved context can be detrimental; only current-anchored retrieval from DINO history surpasses this baseline, reaching $72.8\%$.
Together, these results show that effective intent modeling requires both a current visual-language anchor and a visually stable DINO representation of recent history.
Detailed results for each subset are provided in the supplementary material.

\paragraph{Cross-Branch Attention Visualization.}
Fig.~\ref{fig:cross_branch_attention} visualizes the attention from action queries to the current VAE and DINO tokens in the MoT mixed self-attention.
Across both tasks, action-to-DINO attention aligns more closely with the manipulated objects and interaction regions, whereas action-to-VAE attention is distributed over broader scene areas.
This qualitative pattern suggests that DINO provides task-focused semantic cues complementary to the fine-grained visual dynamics represented by VAE latents.

\section{Conclusion}

Motivated by the robustness limitations exposed by
\textbf{Training-Distribution Hallucination}, we introduced
\textbf{ST-WAM} to improve video-generative World Action Models under visual distribution shifts.
ST-WAM uses DINOv3 in complementary temporal directions:
DSFE complements fine-grained VAE dynamics with visually stable future semantics, while CAIR retrieves task-relevant intent from recent semantic history under the current visual-language context.
Extensive simulation and real-world experiments demonstrate substantially improved robustness while preserving strong in-distribution performance, efficient action-only inference, and freedom from large-scale embodied pretraining.
Overall, our results highlight semantic-temporal modeling as an effective direction for robust world-action learning beyond pixel-centric futures.
Future work will extend ST-WAM beyond visual distribution shifts to changes in physical dynamics and embodiments.

\bibliography{aaai2027}


\end{document}